\documentclass[sigconf]{acmart}
\copyrightyear{2023}
\acmYear{2023}
\setcopyright{rightsretained}
\acmConference[SIGGRAPH '23 Conference Proceedings]{Special Interest Group on Computer Graphics and Interactive Techniques Conference Conference Proceedings}{August 6--10, 2023}{Los Angeles, CA, USA}
\acmBooktitle{Special Interest Group on Computer Graphics and Interactive Techniques Conference Conference Proceedings (SIGGRAPH '23 Conference Proceedings), August 6--10, 2023, Los Angeles, CA, USA}
\acmDOI{10.1145/3588432.3591516}
\acmISBN{979-8-4007-0159-7/23/08}

\usepackage{booktabs} 

\usepackage[ruled]{algorithm2e} 

\SetAlFnt{\small}
\SetAlCapFnt{\small}
\SetAlCapNameFnt{\small}
\SetAlCapHSkip{0pt}

\acmJournal{TOG}

\usepackage{float}

\begin{document}
\title{Nerfstudio: \\ A Modular Framework for Neural Radiance Field Development}

\author{Matthew Tancik}
    \orcid{0000-0002-9059-7909}
    \affiliation{
        \country{ }
    }
    \authornote{Authors contributed equally}
\author{Ethan Weber}
    \orcid{0000-0002-1117-2372}
    \affiliation{
        \country{ }
    }
    \authornotemark[1]
\author{Evonne Ng}
    \orcid{0009-0001-0489-7551}
    \affiliation{
        \country{ }
    }
    \authornotemark[1]
\author{Ruilong Li}
    \orcid{0009-0005-7426-7650}
    \affiliation{
        \country{ }
    }
\author{Brent Yi}
    \orcid{0009-0009-8408-0717}
    \affiliation{
        \country{ }
    }
\author{Justin Kerr}
    \orcid{0000-0002-0536-4853}
    \affiliation{
        \country{ }
    }
\author{Terrance Wang}
    \orcid{0009-0008-5414-4161}
    \affiliation{
        \country{ }
    }
\author{Alexander Kristofferson}
    \orcid{0009-0008-0090-3038}
    \affiliation{
        \country{ }
    }
\author{Jake Austin}
    \orcid{0009-0004-1821-0644}
    \affiliation{
        \country{ }
    }
\author{Kamyar Salahi}
    \orcid{0000-0002-4927-2184}
    \affiliation{
        \country{ }
    }
\author{Abhik Ahuja}
    \orcid{0009-0006-7670-9077}
    \affiliation{
        \country{ }
    }
\author{David McAllister}
    \orcid{0009-0009-0751-2256}
    \affiliation{
        \country{ }
    }
\author{Angjoo Kanazawa}
    \orcid{0000-0003-2592-8430}
    \affiliation{
        \institution{University of California, Berkeley}
        \country{ }
        }


\renewcommand\shortauthors{Tancik*, Weber*, Ng*, et al.}

\begin{abstract}
Neural Radiance Fields (NeRF) are a rapidly growing area of research with wide-ranging applications in  computer vision, graphics, robotics, and more. In order to streamline the development and deployment of NeRF research, we propose a modular PyTorch framework, Nerfstudio. Our framework includes plug-and-play components for implementing NeRF-based methods, which make it easy for researchers and practitioners to incorporate NeRF into their projects. Additionally, the modular design enables support for extensive real-time visualization tools, streamlined pipelines for importing captured in-the-wild data, and tools for exporting to video, point cloud and mesh representations. The modularity of Nerfstudio enables the development of Nerfacto, our method that combines components from recent papers to achieve a balance between speed and quality, while also remaining flexible to future modifications. To promote community-driven development, all associated code and data are made publicly available with open-source licensing.
\end{abstract}

%
%



%
%


\begin{teaserfigure}
\includegraphics[width=\textwidth]{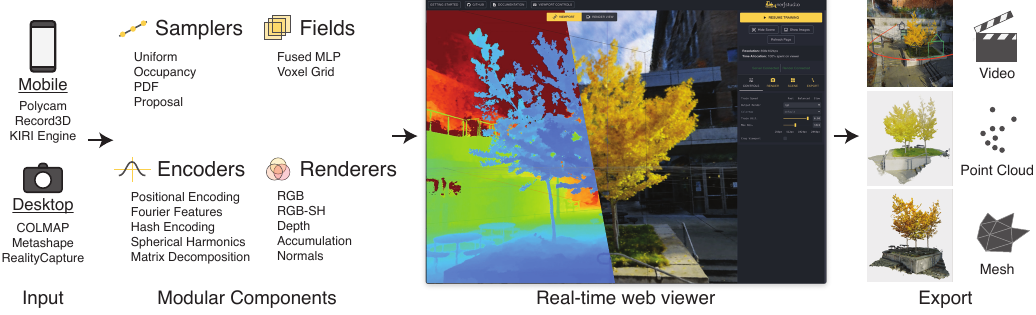}
\caption{\textbf{Nerfstudio framework.} Nerfstudio is a Python framework for Neural Radiance Field (NeRF) development. Nerfstudio supports multiple input data pipelines, is built around multiple modular core NeRF components, integrates with a real-time web viewer, and supports multiple export modalities. The goal of the Nerfstudio framework is to simplify the development of custom NeRF methods, processing of real-world data, and interacting with reconstructions.}
\label{fig:teaser}
\end{teaserfigure}

\maketitle

\section{Introduction}

Neural Radiance Fields (NeRFs)~\cite{mildenhall2021nerf} are gaining popularity for their ability to create 3D reconstructions in real-world settings, with rapid research in the area pushing the field forward.
Since the introduction of NeRFs in 2020, there has been an influx of papers focusing on advancements to the core method including few-image training~\cite{yu2021pixelnerf__pixelnerf,wang2021ibrnet}, explicit features
for editing~\cite{liu2020neural__nsvf, wang2022clip, zhang2022arf}, 
surface representations for high-quality 3D mesh exports~\cite{oechsle2021unisurf, yariv2021volume, wang2021neus}, 
speed improvements for real-time rendering and training~\cite{yu2021plenoxels__plenoxels, sun2022direct__dvgo, muller2022instant}, 3D object generation~\cite{poole2022dreamfusion}, and more~\cite{xie2022neural}. 


These research innovations have driven interests in a wide variety of disciplines in both academia and industry. 
Roboticists have explored using NeRFs for manipulation, motion planning, simulation, and mapping~\cite{kerrevo__evonerf,adamkiewicz2021,driess2022compNerfPreprint,byravan2022nerf2real,Zhu2022CVPR,simeonov2021neural}. NeRFs are also explored for tomography applications~\cite{ruckert2022neat}, as well as perceiving people in videos~\cite{pavlakos2022one}. Visual effects and gaming studios are exploring the technology for production and digital asset creation. 
News outlets capture NeRF portraits to tell stories in new formats~\cite{nytnerf}.
The potential applications are vast, and even startups~\footnote{https://lumalabs.ai/} are emerging to focus on deploying this technology.

Despite the growing use of NeRFs, support for development is still rudimentary. Due to the influx of papers and lack of code consolidation, tracking progress is difficult. Many papers implement features in their own siloed repository. This complicates the process of transferring features and research contributions across different implementations. Additionally, few tools exist to easily run NeRFs on real-world data collected by users. 
To address these challenges, 
we present Nerfstudio (Fig.~\ref{fig:teaser}), a modular framework that consolidates NeRF research innovations and makes them easier to use in real-world applications.

Furthermore, while NeRFs solve an inherently visual task, there is a lack of comprehensive and extensible tools for visualizing and interacting with NeRFs trained on real-world data. Despite the availability of several NeRF repositories, existing implementations are often focused on achieving state-of-the-art results on metrics such as PSNR, SSIM, and LPIPS. These evaluations are typically based on held-out images along the capture trajectory that are similar to the training images. This often makes them misleading indicators of performance for many real-world applications when data is captured in unstructured environments and novel views are rendered with large baselines. Qualitative evaluations have historically been a challenge due to the computational demands of NeRF, which often resulted in rendering times up to multiple seconds per image. Recent developments such as Instant-NGP~\cite{muller2022instant} significantly reduce computational overhead, enabling real-time training and rendering. However, Instant-NGP relies significantly on GPU acceleration with custom CUDA kernels, making development and quick prototyping a challenge. We present a framework that enables interactive visualizations while also being flexible and model-agnostic.  

\begin{figure}[t]
\includegraphics[width=0.47\textwidth]{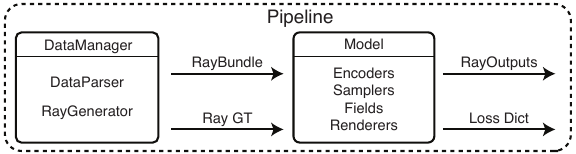}
\caption{\textbf{Pipeline components.} Each NeRF method is implemented as a custom Pipeline. DataManagers process input images into bundles of rays (RayBundles) that get rendered by the Model to produce a set of NeRF outputs (RayOutputs). A dictionary of losses supervises the pipeline end-to-end.}
\label{fig:pipeline}
\vspace{-5mm}
\end{figure}

Nerfstudio is an extensible and versatile framework for neural radiance field development. Our design goals are the following:
\begin{enumerate}
    \item Consolidating various NeRF techniques into reusable, modular components.
    \item Enabling real-time visualization of NeRF scenes with a rich suite of controls.
    \item Providing an end-to-end, easy-to-use workflow for creating NeRFs from user-captured data.
\end{enumerate}

For modularity, we devise an organization among components across various NeRFs that allows abstracting away method-specific implementations. 
Our real-time visualizer is designed to work with any model during training or testing. Furthermore, the visualizer is hosted on the web, making it accessible without requiring a local GPU machine. 
The modular nature of our framework facilitates the integration of various data input formats, thereby simplifying the workflow for incorporating user-captured real-world scenes. We provide support for images and videos with various camera types, as well as other mobile capture applications (Polycam, Record3D, KIRI Engine) and outputs from popular photogrammetry software like RealityCapture and Metashape. In particular, integration with these applications enable users to by-pass structure-from-motion tools like COLMAP~\cite{schoenberger2016sfm}, which can be time-consuming. Furthermore, we provide support for multiple export formats, including video, depth maps, point clouds, and meshes.

\begin{figure*}[t]
\includegraphics[width=\textwidth]{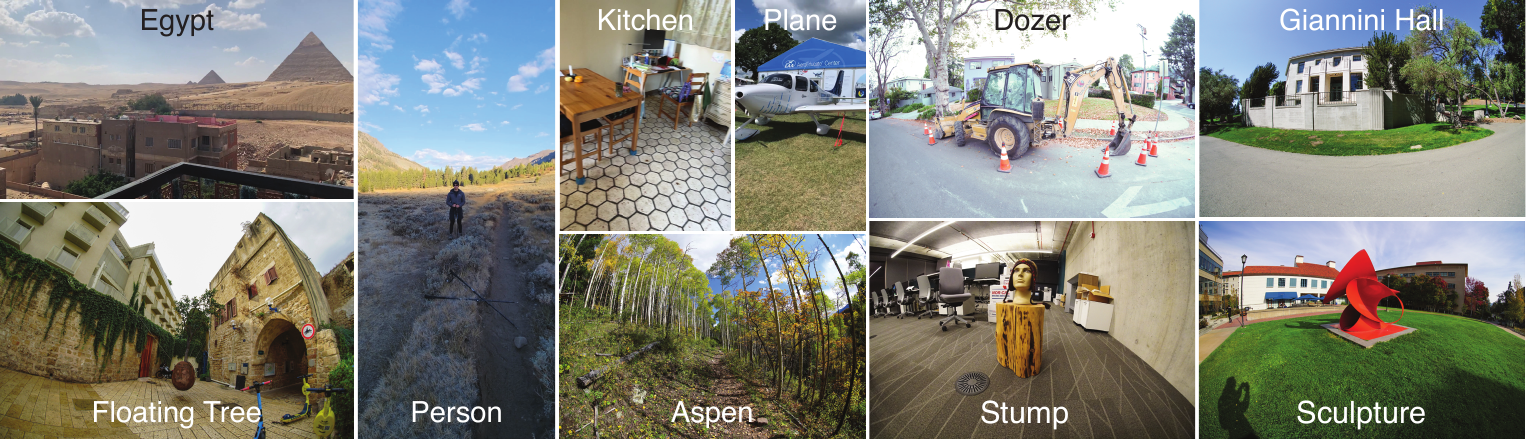}
\caption{\textbf{Nerfstudio Dataset.} Our Nerfstudio Dataset contains 10 scenes: 4 phone captures with pinhole lenses and 6 mirrorless camera captures with a fisheye lens. We focus our efforts on real-world data, and these scenes can help benchmark progress.}
\label{fig:nerfstudio_dataset}
\end{figure*}

The modularity of Nerfstudio enables developing Nerfacto, our method that combines components from recent papers to achieve a balance between speed and quality. We show that this method is comparable to the other state-of-the-art methods such as MipNeRF-360~\cite{barron2022mip} while achieving an order of magnitude speedup. We also conduct an ablation study that demonstrates its flexibility on a new in-the-wild dataset consisting of 10 in-the-wild scenes. Our findings highlight the limitations of commonly used NeRF metrics and the importance of a real-time viewer for qualitative assessments. The potential of our framework as a consolidated codebase for NeRF research is reflected in the traction thus far with extensions such as SDFStudio~\cite{Yu2022SDFStudio}. Furthermore, Nerfstudio is an open-source project with active improvements from both academic and industry contributors.

\section{Related Works}

\subsection{Frameworks and tools}

Software frameworks have played a crucial role in consolidating and driving the advancement of various fields. In deep learning, Caffe~\cite{jia2014caffe}, TensorFlow~\cite{abadi2016tensorflow}, and PyTorch~\cite{paszke2019pytorch} provide readily usable machine learning functionalities. Similarly, frameworks such as PyTorch3D~\cite{ravi2020accelerating__pytorch3d} and Kornia~\cite{eriba2019kornia} provide reusable components for 3D computer vision tasks. Other examples of frameworks include Mitsuba3~\cite{Mitsuba3__mitsuba}, Halide~\cite{ragan2013halide}, Taichi~\cite{hu2019taichi}, and Reyes~\cite{cook1987reyes} for graphics, Phototourism~\cite{snavely2006photo} and COLMAP~\cite{schoenberger2016sfm, schoenberger2016mvs, schoenberger2016vote} for photogrammetry and visualization, and AverageExplorer~\cite{zhu2014averageexplorer} for data collection. Despite the diversity of topics covered, each of these frameworks originated from the need to provide reusability and reproducibility to a rapidly expanding field. In light of the fast-paced growth of NeRFs in both academia and industry, Nerfstudio aims to streamline advancements in neural rendering by offering a flexible and comprehensive framework for development.

\subsection{NeRF codebases}

In recent years, several codebases for NeRFs have gained popularity among the research community, including the original NeRF codebase\cite{mildenhall2021nerf}, nerf-pytorch~\cite{lin2020nerfpytorch,krishna_nerf-pytorch}, Nerf\_pl~\cite{queianchen_nerf__nerfpl}, Instant NGP~\cite{muller2022instant}, torch-ngp, Ngp\_pl, and MultiNeRF~\cite{multinerf2022}. 
Due to the lack of consolidation, there exists a significant number of NeRF repositories that focus on improving specific components of specific algorithms. For example, Mip-NeRF~\cite{barron2021mip} aims to address the anti-aliasing problem of NeRF~\cite{mildenhall2021nerf} and Mip-NeRF 360~\cite{barron2022mip} addresses the limitations of Mip-NeRF. Additionally, Plenoxels~\cite{yu2021plenoxels__plenoxels}, KiloNeRF \cite{Reiser_2021_ICCV}, SNerg \cite{hedman2021snerg}, FastNerf \cite{9710021}, DONeRF \cite{neff2021donerf}, TensoRF~\cite{chen2022tensorf_tensorf} and InstantNGP~\cite{muller2022instant} propose different approaches to address the problem of computational efficiency. Furthermore, RawNeRF~\cite{mildenhall2022nerf_rawnerf}, Ref-NeRF~\cite{verbin2022ref_refnerf}, and NeRF-W~\cite{martin2021nerf__nerfw} each address distinct challenges related to NeRF, resulting in parallel, non-interacting implementations. Nerfstudio aims to address the lack of consolidated development in the field of NeRFs by consolidating critical techniques introduced in the existing literature. This allows for more efficient and effective experimentation with combining components from multiple solutions into a single, comprehensive method, and facilitates the ability of the community to build upon existing prior approaches.

\subsection{Neural rendering frameworks}

Concurrent efforts such as NeRF-Factory~\cite{nerf_factory}, NerfAcc~\cite{li2022nerfacc}, MultiNeRF~\cite{multinerf2022}, and Kaolin-Wisp~\cite{KaolinWispLibrary} all make significant efforts in advancing the usability of NeRFs. While NeRF-Factory consolidates multiple prior works into a single repository, it places less emphasis on reusable modules shared across these prior works and focuses more on benchmarking. NerfAcc prioritizes pythonic modularity, but focuses primarily on the lower-level components rather than the entire pipeline. Kaolin-Wisp and Multi-NeRF each consolidate multiple paper implementations into a single repository. None of these repositories are as comprehensive as Nerfstudio in delivering our three design goals: modularity, real-time visualization, and end-to-end usability for user-captured data. Furthermore, Nerfstudio is released under an Apache2 license, which allows for its use by both researchers and companies.

\section{Framework Design}

The goals of Nerfstudio are to provide (1) modularity, (2) real-time visualization for development, and (3) ease of use with real data.
In designing the framework, we consider trade-offs against designs that optimize for faster rendering or higher quality results on synthetic scenes. 
For instance, we prefer an implementation that allows for a modularized pythonic non-CUDA method (i.e., where CUDA functionality is exposed via a PyTorch API) over one that supports a faster, non-modularized CUDA method (where CUDA code is written directly).
Additionally, our design choices lead to simpler interfacing with an extensive visualization ecosystem which supports real-time rendering during test and train with custom camera paths. Finally, we focus on delivering results for real-world data rather than synthetic scenes to address audiences outside research including those in industry and non-technical users.

With these three goals, the design of Nerfstudio promotes collaborations by providing a consolidated platform on which people can request for or contribute to new features. The long-term goal is for Nerfstudio to continue improving through community-driven contributions.

\subsection{Modularity}

We propose an organization of components that is both intuitive and abstract, enabling the implementation of existing and novel NeRFs by swapping reusable components.
Fig.~\ref{fig:teaser} shows a subset of the components types and implementations we currently have available in Nerfstudio.

\subsection{Visualization for development}

The Nerfstudio real-time viewer offers an interactive and intuitive way to visualize Neural Radiance Fields (NeRFs) during both training and testing phases. To ensure ease of use, the visualizer is simple to install, works seamlessly across both local and remote GPU compute environments, supports different models, and offers a user interface for creating and rendering custom camera paths, shown in Fig.~\ref{fig:geometry_export} (a).

Our real-time visualization interface is particularly useful for qualitatively evaluating a model, allowing for more informed decisions during method development.
While metrics such as PSNR can provide some insight, they do not offer a comprehensive understanding of performance--especially for views that are far away from the capture trajectory. Qualitative evaluation with an interactive viewer addresses these limitations and allows developers to gain a more holistic understanding of the model performance.

\subsection{Easy workflow for user-captured data}

While we offer support for synthetic datasets (Blender~\cite{mildenhall2021nerf}, D-NeRF~\cite{pumarola2021d}), in Nerfstudio we focus primarily on "real world data" --- images or videos from a physical phone or camera. To this end, we present a new Nerfstudio Dataset (shown in Fig.~\ref{fig:nerfstudio_dataset}) composed of real-world scenes casually captured with mobile phones and a mirrorless camera. Our motivation is to provide a framework compatible with a diverse array of applications which requires supporting real data. For instance, a few use cases for Nerfstudio outside of research include VFX, gaming, and non-technical film-makers who create 3D and video art. To support this wide range of expertise in NeRFs, we ensure our codebase is easily installable and deployable.

\section{Core components}

The proposed framework of Nerfstudio, illustrated in Fig.~\ref{fig:pipeline}, is based on the conceptual grouping of NeRF methods into a series of basic building blocks. Nerfstudio takes a set of posed images and optimizes for a 3D representation of the scene, which is defined by radiance (color), density (structure), and possibly other quantities (semantics, normals, features, etc.). 
We ingest these inputs into the framework which comprises of a DataManager and a Model, where the DataManager is responsible for (1) parsing image formats via a DataParser and (2) generating rays as RayBundles. These rays are then passed into a Model, which will query Fields and render quantities. Finally, the whole Pipeline is supervised end-to-end with a loss.

\subsection{DataManagers and DataParsers}

The first step of the Pipeline is the DataManager which is responsible for turning posed images into RayBundles, which are slices of 3D space that start at a camera origin. Within the DataManager, the DataParser first loads the input images and camera data.
The DataParser is designed to be compatible with arbitrary data formats such as COLMAP.
Previous research codebases primarily utilize COLMAP with helper scripts~\cite{muller2022instant}, however, COLMAP can be challenging to install and use for non-technical users.
To make the framework more accessible to a wider range of users, including scientists, artists, photographers, hobbyists and journalists, we have implemented DataParsers for mobile apps (Record3D, Polycam, KIRI Engine) and 3D tools such as Metashape and Reality Capture.
Once the images are properly loaded and formatted, the DataManager iterates through the data, generating RayBundles and ground truth supervision. It can also optimize camera poses during training.

\subsection{RayBundles, RaySamples, and Frustums}
\begin{figure}[ht]
\includegraphics[width=0.47\textwidth]{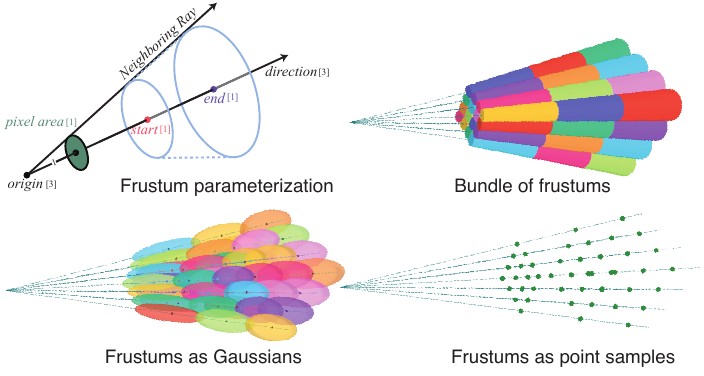}
\caption{\textbf{Sample representations.} (Top) We define a frustum as a cone with a start and end. This region of space can be converted into Gaussians (bottom left) or point samples (bottom right) depending on the field input format.}
\label{fig:sample_representations}
\vspace{-5mm}
\end{figure}

NeRFs operate on regions of 3D space, which can be parametrized in many different ways. We have adopted a more generic representation of 3D space through the use of Frustum for both point-based and volume-based samples. The RayBundles, which are primitives that represent a slice through 3D space, are parameterized with an origin, direction, and other meta-information such as camera indices and time. By specifying the interval bin spacing, the RayBundles generate RaySamples, which represent sampled chunks of 3D space along each ray. These chunks, represented as Frustums, can be encoded either as point samples~\cite{mildenhall2021nerf} or as Gaussians with mean and covariance~\cite{barron2021mip}, which have been shown to help with anti-aliasing. This abstraction allows for flexibility in representation, as the user can decide which representation to use with a simple function call. A visualization of this abstraction can be found in Fig.~\ref{fig:sample_representations}.

\subsection{Models and Fields}
The RayBundles are sent to Models as input, which samples them into RaySamples. The RaySamples are consumed by Fields to turn regions of space (i.e., Frustums) into quantities such as color or density.
The Nerfstudio framework contains various implementations of models and fields. 
We've implemented various feature encoding schemes including fourier features, hash encodings~\cite{muller2022instant}, spherical harmonics, and matrix decompositions~\cite{chen2022tensorf_tensorf}. Field components include fused MLPs, voxel grids, and surface normal MLPs~\cite{verbin2022ref_refnerf}, activation functions, spatial distortions~\cite{barron2022mip}, and temporal distortions~\cite{pumarola2021d}.

\subsection{Real-time Web Viewer}
We draw inspiration from the real-time viewer presented in Instant NGP~\cite{muller2022instant}, which facilitates real-time rendering during training. However, the viewer in Instant NGP is designed to work on local compute, which can be cumbersome to setup in remote settings. To address this issue, we have developed a ReactJS-based web viewer packaged as a publicly hosted website.

\begin{figure}[ht]
\includegraphics[width=0.47\textwidth]{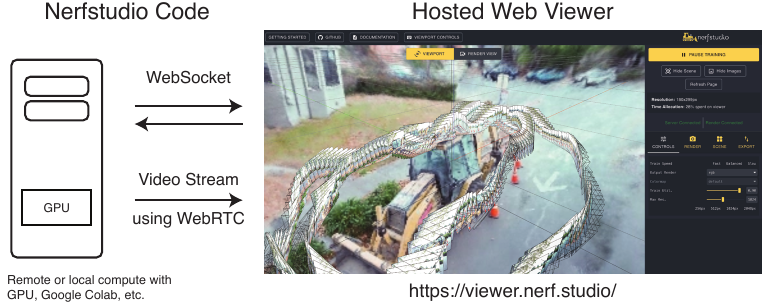}
\caption{\textbf{Web viewer design.} A machine with a GPU (left) starts a NeRF training session. When a user navigates to to the hosted web viewer (right), the viewer client will establish WebSocket and WebRTC connections with the training session.}
\label{fig:viewer}
\vspace{-3mm}
\end{figure}

The viewer is designed to be accessible to a wide range of users, including those utilizing both local and remote GPUs. The process of utilizing remote compute is streamlined, requiring only the forwarding of a port locally via SSH. Once training begins, the web interface renders the NeRF in real-time as training progresses (See Fig.~\ref{fig:figure_page_1}). Users can pan, zoom and rotate around the scene as the optimization runs or while evaluating a trained model. The design of the viewer is illustrated in Fig.~\ref{fig:viewer}.

\subsubsection{Implementation}

Real-time training visualization utilizes WebSockets and WebRTC to establish a connection between the NeRF training session and the web client. This approach eliminates the need to install 
local screens and other GUI software.
Upon opening the web viewer, a WebSocket connection is established with the training session, which subsequently populates the scene with training images as illustrated in Fig.~\ref{fig:viewer} (right). The web viewer continuously streams the viewport camera pose to the training session during the training process. The training session utilizes this camera pose to render images and transmits them via a WebRTC video stream. Additionally, the viewer camera controls and UI are implemented using ThreeJS, allowing us to overlay 3D assets such as images, splines, and cropping boxes in front of the NeRF renderings. For instance, the viewer displays training images at their capture locations, letting users intuitively compare performance at seen and novel viewpoints.  

\subsubsection{Viewer features.}

\begin{figure*}[t]
\includegraphics[width=\textwidth]{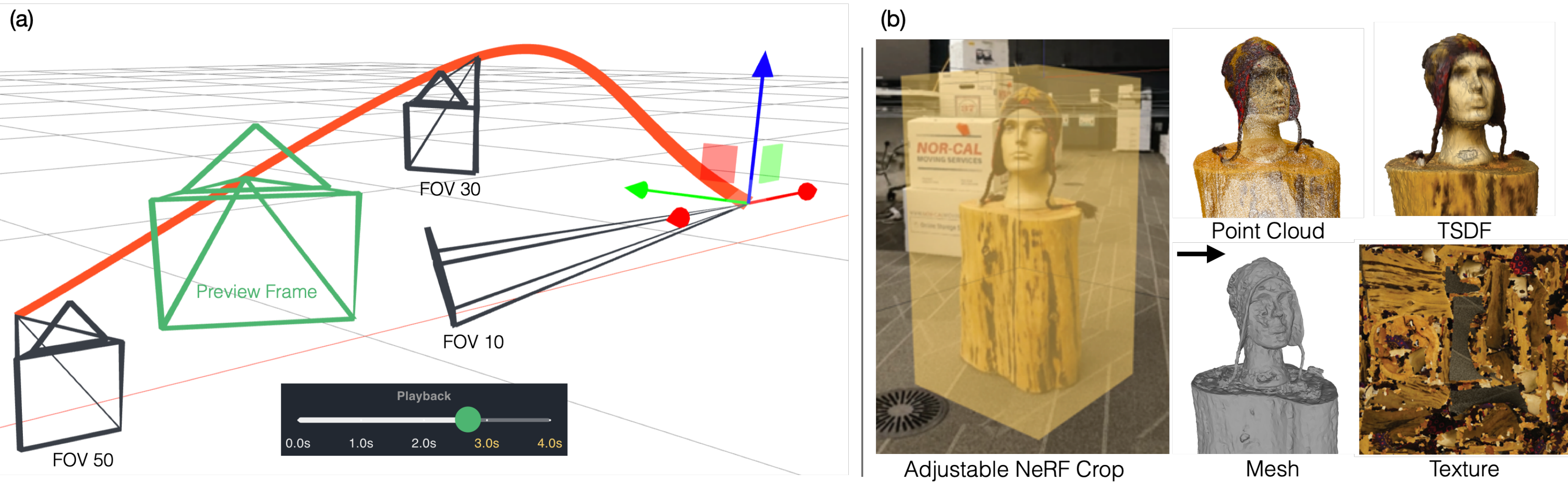}
\caption{\textbf{Exporting videos and geometry.} We make exporting videos (a) and geometry (b) easy with real-data captures. The left side shows the interactive camera trajectory editor, which allows animatable poses, FOVs, and speed, to eventually render videos of NeRF's outputs.  On the right we show the cropping interface in the viewer and resulting export formats including point clouds, TSDFs, and textured meshes.}
\label{fig:geometry_export}
\end{figure*}

Our viewer is compatible with different models of varying rendering speeds. We accomplish this by balancing the computation of training and viewer rendering on a single GPU. Similar to Instant-NGP~\cite{muller2022instant}, we adjust the rendering resolution based on the speed of the camera movement. When the camera moves quickly, the rendering resolution will be smaller to maintain a frame rate and prevent lag in the user experience. We can also reduce the time spent on training and allocate more resources for rendering in the viewer. Some of the features of our viewer include:

\begin{itemize}
  \item Switching between various model outputs (e.g., rgb, depth, normals, semantics).
  \item Creating custom camera paths composed of keyframes with position and focal length interpolation (Fig.~\ref{fig:geometry_export}).
  \item Visualizing the captured training images in 3D.
  \item Crop and export options for point clouds and meshes.
  \item Mouse and keyboard controls to easily navigate in the scene.
\end{itemize}

The viewer played an instrumental role in providing qualitative assessments that informed design choices in our
default method Nerfacto.
Other codebases have integrated our viewer into their own codebases, including ArcNerf~\cite{tencent_arcnerf} and SDFStudio~\cite{Yu2022SDFStudio}.

\subsection{Geometry Export}
Many creators and artists have workflows that require exporting to point clouds or meshes for further processing and incorporation in downstream tools such as game engines. Hence, our framework accommodates various export methods and facilitates the easy addition of new export methods. Fig.~\ref{fig:geometry_export}b illustrates our export interface, as well as some of the supported formats, including point clouds, a truncated signed distance function (TSDF) to mesh, and Poisson surface reconstruction~\cite{kazhdan2006poisson}. We apply texture to the mesh by densely sampling the texture image, utilizing barycentric interpolation to determine corresponding 3D point locations, and rendering short rays near the surface along the normals to obtain RGB values.

\section{Nerfacto Method}

\begin{figure}[ht]
\includegraphics[width=0.47\textwidth]{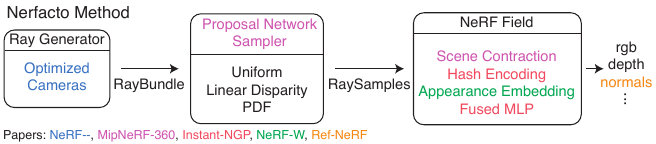}
\caption{\textbf{Nerfacto method.} Diagram of the Nerfacto method. It combines features from many papers (bottom left). The method will evolve over time as new papers and features are added to the Nerfstudio codebase.}
\label{fig:nerfacto}
\vspace{-5mm}
\end{figure}

We leverage our modular design to integrate ideas from multiple research papers into our default and recommended method, Nerfacto. This method is heavily influenced by the structure of MipNeRF-360~\cite{barron2022mip}, but certain parts of the original design are replaced to improve performance. We reference papers such as NeRF-{}-~\cite{wang2021nerf}, Instant-NGP~\cite{muller2022instant}, NeRF-W~\cite{martin2021nerf__nerfw}, and Ref-NeRF~\cite{verbin2022ref_refnerf} in Nerfacto. Fig.~\ref{fig:nerfacto} illustrates how these papers are used. 

\subsection{Ray generation and sampling} The Nerfacto method first optimizes camera views using an optimized SE(3) transformation~\cite{wang2021nerf, lin2021barf, tancik2022blocknerf}. These camera views are then used to generate RayBundles. To improve the efficiency and effectiveness of the sampling process, we employ a piece-wise sampler. This sampler samples uniformly up to a fixed distance from the camera, followed by samples that are distributed such that the step size increases with each sample. 
This allows efficient sampling of distant objects while still maintaining a dense set of samples for nearby objects. These samples are then fed into a proposal network sampler, proposed in the MipNeRF-360 method~\cite{barron2022mip}.
The proposal sampler consolidates the sample locations into regions of the scene that contribute most to the final render, typically the first surface intersection. This importance sampling greatly improves reconstruction quality. Furthermore, we use a small fused MLP with a hash encoding~\cite{muller2022instant} for the scene's density function as it has been found to have sufficient accuracy and is computationally efficient. 
To further reduce the number of samples along rays, the proposal network sampler can contain multiple density fields. These density fields iteratively reduce the number of samples. Empirically, using two density fields works well. In our base Nerfacto configuration, we generate 256 samples from the piece-wise sampler, which gets resampled into 96 samples in the first iteration of the proposal sampler followed by 48 samples in the second.

\subsection{Scene contraction and NeRF field}
\begin{figure}[ht]
\includegraphics[width=0.47\textwidth]{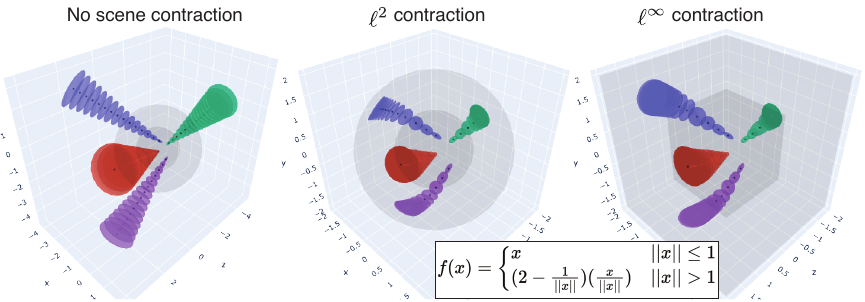}
\caption{\textbf{Scene contraction.} Here we show cameras contained in an inner sphere with Gaussian samples along rays. Scene contraction warps the unbounded samples into bounded space before querying a NeRF field. We use $L^{\infty}$ contraction rather than MipNeRF-360's $L^{2}$ contraction to better accommodate the geometry/capacity of the hash grid.}
\label{fig:scene_contraction}
\end{figure}

Many real-world scenes are unbounded, meaning they could extend indefinitely. This poses a challenge for processing as input samples could have position values that vary across many scales of magnitude. To overcome this issue, we utilize \textit{scene contraction}, which compresses the infinite space into a fixed-size bounding box. Our method of contraction is based on the one proposed in MipNeRF-360~\cite{barron2022mip}, but we use $L^{\infty}$ norm contraction instead of $L^{2}$ norm, which contracts to a cube rather than a sphere. The cube better aligns with voxel-based hash encodings. Fig.~\ref{fig:scene_contraction} illustrates how $L^{\infty}$ contraction maps samples into the range with minimum values of {-2,-2,-2} and maximum values of {2,2,2}. These samples can then be used with the hash encoding introduced by Instant-NGP and is available via the tiny-cuda-nn~\cite{tiny-cuda-nn} Python bindings.

Nerfacto's field incorporates per-image appearance embeddings to account for differences in exposure among training cameras \cite{martin2021nerf__nerfw}. Additionally, we use techniques from Ref-NeRF \cite{verbin2022ref_refnerf} to compute and predict normals. Nerfacto is implemented using PyTorch, which allows for easy customization and eliminates the need for complex and custom CUDA code. We will incorporate new papers into Nerfacto as the field progresses.

\begin{figure*}[t]
\includegraphics[width=\textwidth]{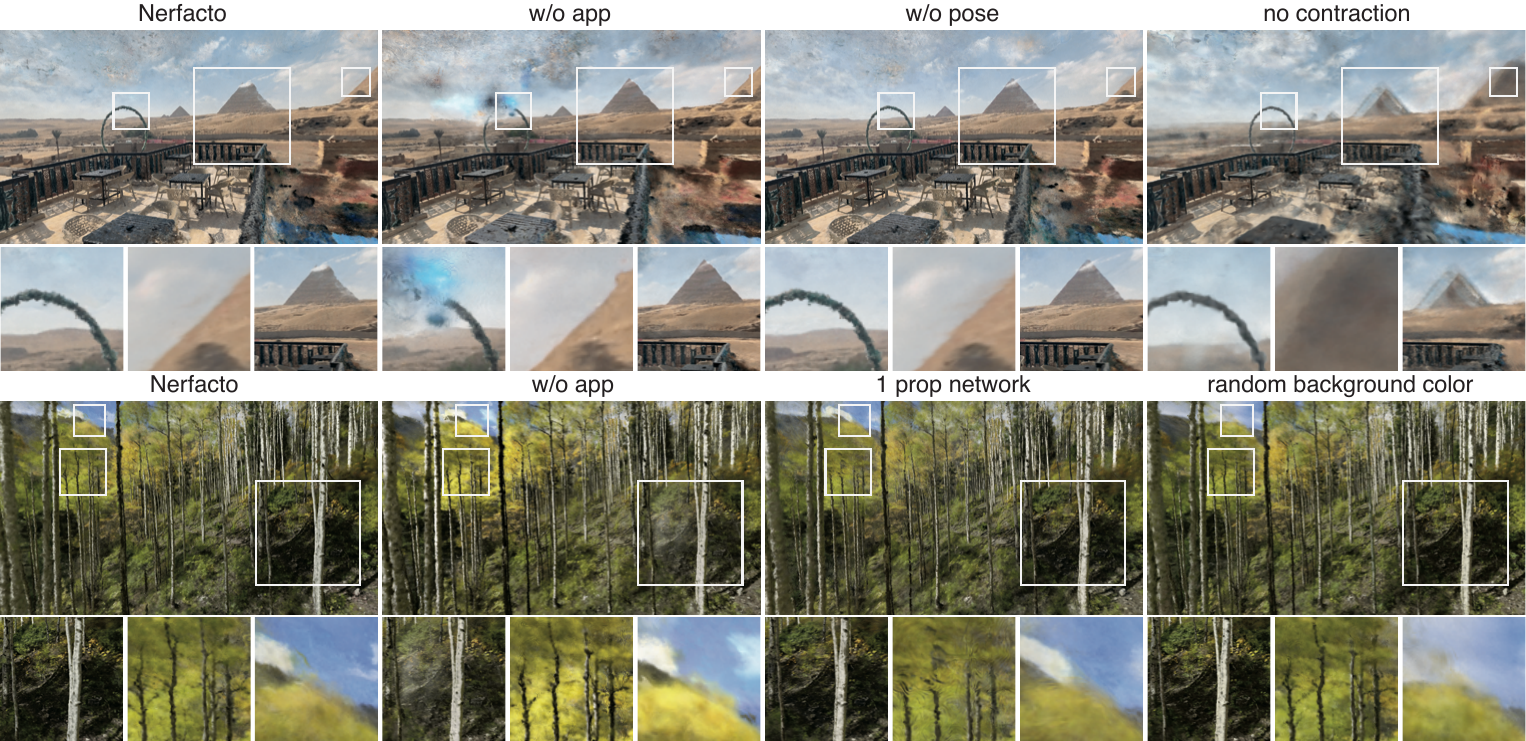}
\caption{\textbf{Nerfstudio ablation qualitative examples.} Here we show renderings from different Nerfacto ablation variants. (Top) is the "Egypt" capture and (bottom) is the "aspen" capture from the Nerfstudio Dataset. These novel views are far from the training images to get a sense of how well these methods perform qualitatively. We zoom in on crops to highlight differences in the rendered images.}
\label{fig:qualitative_nerfacto_examples}
\end{figure*}

\section{Nerfstudio Dataset}
Our "Nerfstudio Dataset" includes 10 in-the-wild captures obtained using either a mobile phone or a mirror-less camera with a fisheye lens. We processed the data using either COLMAP or the Polycam app to obtain camera poses and intrinsic parameters. Our goal is to provide researchers with more 360 real-world captures that are not limited to forward-facing scenes~\cite{mildenhall2019local__llff}. Our dataset is similar to MipNeRF-360~\cite{barron2022mip} but does not focus on a central object and includes captures with varying degrees quality. We have used this dataset to select the default settings for our proposed NeRF-based method, Nerfacto, and we encourage other researchers to similarly employ real-world data in the development and evaluation of NeRF methods.

\section{Experiments}

We benchmark Nerfacto against a state-of-the-art method MipNeRF-360 and emphasize the modularity of our repository by conducting ablation studies. Furthermore, we highlight the limitations of commonly used evaluation metrics such as PSNR, SSIM, and LPIPS when applied to subsampled evaluation images.

\subsection{Mip-NeRF 360 Dataset Comparison}

Here we compare Nerfacto with numbers reported in the MipNeRF-360~\cite{barron2022mip} paper. We evaluate on their 7 publicly available scenes. We train our method for up to 70K iterations ($\sim$30 minutes) on an NVIDIA RTX A5000, but we also report results at 5K iterations ($\sim$2 minutes).

\subsubsection{Evaluation protocol.} The evaluation protocol followed is similar to that of MipNeRF360, but we process their data using our COLMAP pipeline to recover poses. The original images were downsampled by a factor of 4x. We used 7/8 of the images for training and the remaining 1/8 images were evenly spaced and used for evaluation. Note that this protocol does not include camera pose optimization as it is not an option implemented in MipNeRF360. 

\subsubsection{Findings.} 
\begin{figure*}[hbtp]
\includegraphics[width=\textwidth]{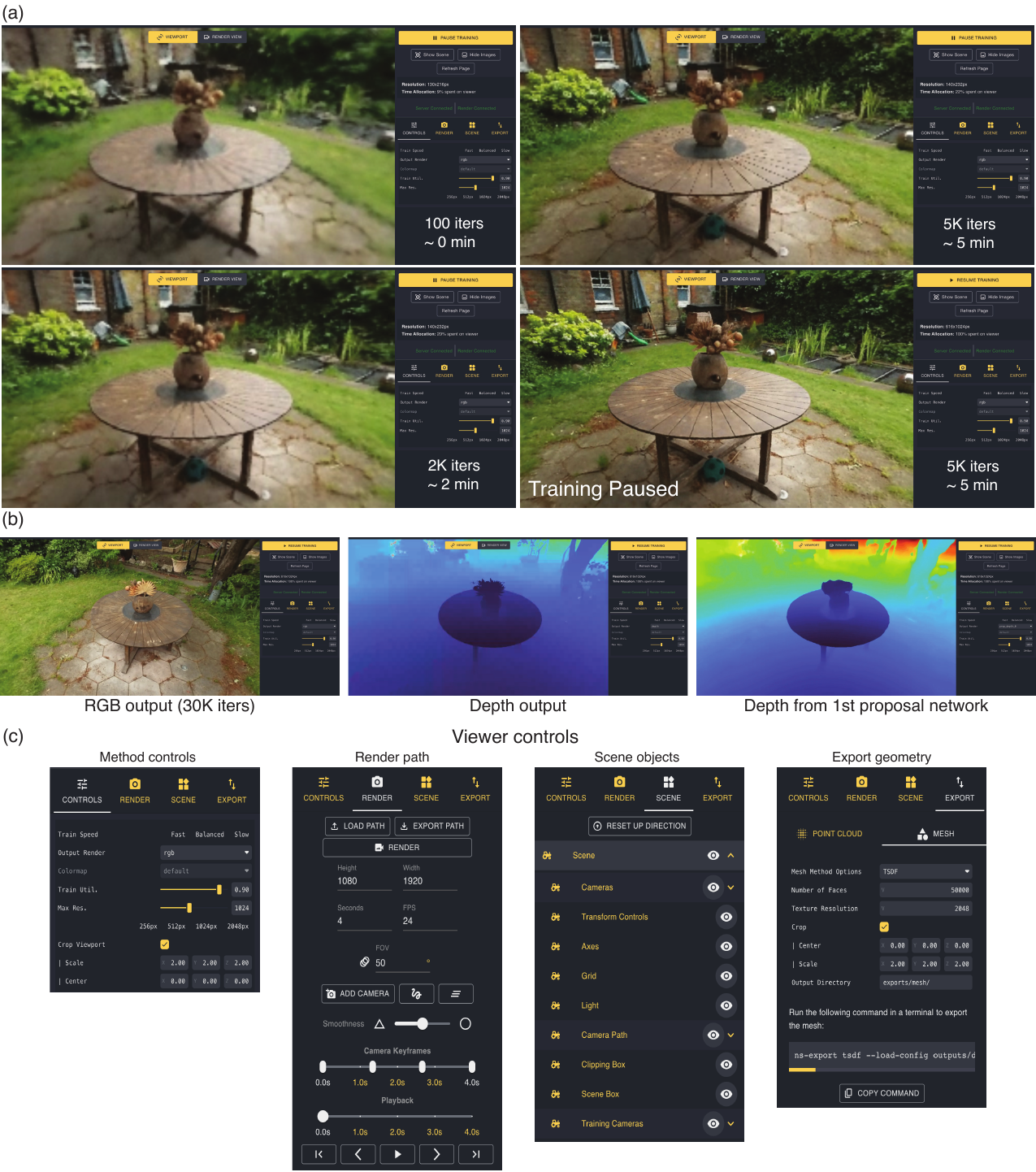}
\caption{\textbf{Real-time viewer use.} (a) Training Nerfacto on the MipNeRF-360 garden scene. Good quality can be achieved after a few minutes. Pausing the training increases the rendered resolution. (b) Visualizing different model outputs with the viewer. (c) Viewer controls and settings available in the viewer.}
\label{fig:figure_page_1}
\end{figure*}

Table~\ref{fig:table-mipnerf-ablations-average} presents the averages of the results across the 7 captures in the MipNeRF-360 dataset. The complete table can be found in the supplementary material. In as little as 5K iterations ($\sim$2 minutes), our Nerfacto method achieves reasonable quality in contrast to MipNeRF-360 which takes several hours on a TPU with 32 cores. Training for up to 70K ($\sim$30 minutes) iterations further improves quality.
While Nerfacto falls short of metric results obtained by MipNeRF-360, we prioritize efficiency and general usability over optimizing quantitative metrics on this particular benchmark. Fig.~\ref{fig:figure_page_1} shows qualitative results on the "garden" scene in our viewer after only a few minutes.

\begin{table}[h]
\caption{\textbf{Average metrics on the MipNeRF360 dataset.} Our methods are evaluated without pose optimization or per-image appearance embeddings. MipNeRF-360 takes several hours to train. Our metrics reported as \{ after 70K iterations ($\sim$30 min) / after 5k iterations ($\sim$2 min) \}.}
\begin{tabular}{l|ccc}
\toprule
Method & $\textbf{PSNR}$ $\uparrow$ & $\textbf{SSIM}$  $\uparrow$ & $\textbf{LPIPS}$ $\downarrow$ \\ 
\midrule
NeRF & 24.85 & 0.659 & 0.426 \\
MipNeRF & 25.12 & 0.672 & 0.414 \\
NeRF++ & 26.21 & 0.729 & 0.348 \\
MipNeRF (big MLP) & 27.60 & 0.806 & 0.251 \\
NeRF++ (big MLP) & 27.66 & 0.803 & 0.265 \\
MipNeRF-360 & 29.23 & 0.844 & 0.207 \\
Nerfacto (ours) & 27.98 / 25.38 & 0.800 / 0.688 & 0.291 / 0.390 \\
\bottomrule
\end{tabular}
\label{fig:table-mipnerf-ablations-average}
\vspace{-5mm}
\end{table}

It is worth emphasizing that our Nerfacto method is optimized for qualitative novel-view quality by using the web viewer, rather than solely relying on common metrics. For further illustration, we refer the reader to the supplementary material where we provide rendered videos from our Nerfacto method.

\subsection{Nerfacto Component Ablations}
\label{sec:nerfacto_ablations}

Given the modularity of our codebase, we can easily conduct ablation studies on our method Nerfacto, a unified approach that combines important components from various papers to achieve a fast, high-quality method. 
We experiment with disabling the pose optimization, appearance embeddings, scene contraction, and variations of the proposal networks, and more. The modularity of our codebase allows for easy implementation of these modifications through the use of different flags with the command line interface.

\subsubsection{Evaluation protocol.} In our ablation study, we utilize the Nerfstudio Dataset for evaluation. Due to the complexity of the appearance embeddings and pose optimization modules, we adopt a test-time optimization procedure for the evaluation. Specifically, we employ Adam optimizers to optimize the evaluation camera poses. Once the camera poses are fixed, we randomly select either the left or right side of the evaluation image and optimize the appearance code as done in Martin et al.~\cite{martin2021nerf__nerfw}. Finally, with the optimized camera pose and appearance embedding, we compute PSNR, SSIM, and LPIPS. For these experiments, 10\% of the images chosen at equal intervals were used for evaluation. This is similar to MipNeRF-360~\cite{barron2022mip} but instead of 1 in every 8 consecutive frames, we use 1 in every $N * D$ where $N$ is the percentage and $D$ is the number of images in the captured dataset. We use $N = 0.1$. We will release this evaluation protocol so future work can run similar experiments.

\begin{table}[h]
\caption{\textbf{Average metrics for ablations on the Nerfstudio Dataset.} We remove and change various components of the Nerfacto method and report \{ PSNR, SSIM, LPIPS \} on the Nerfstudio Dataset. Further details on the experiments can be found in the supplementary material.}
\begin{tabular}{l|ccc}
\toprule
Nerfacto method & $\textbf{PSNR}$ $\uparrow$ & $\textbf{SSIM}$  $\uparrow$ & $\textbf{LPIPS}$ $\downarrow$ \\
\midrule
Nerfacto (default) & 20.99 & 0.663 & 0.389 \\
w/o pose & 20.93 & 0.659 & 0.393 \\
w/o app & 22.65 & 0.672 & 0.406 \\
w/o pose  \& app & 22.53 & 0.671 & 0.411 \\
1 prop network & 21.07 & 0.669 & 0.396 \\
l2 contraction & 20.98 & 0.664 & 0.388 \\
shared prop network & 20.95 & 0.661 & 0.391 \\
random backg. color & 21.00 & 0.663 & 0.392 \\
no contraction & 18.59 & 0.534 & 0.506 \\
synthetic on real & 20.09 & 0.542 & 0.509 \\
\bottomrule
\end{tabular}
\label{fig:table-nerfacto-ablations-average}
\vspace{-5mm}
\end{table}

\subsubsection{Findings.} Table~\ref{fig:table-nerfacto-ablations-average} presents the average results of our ablation studies. The complete table for all 10 scenes can be found in the supplementary material. 
This study highlights the challenge in extracting meaningful insights from quantitative metrics alone (Table~\ref{fig:table-nerfacto-ablations-average}), due to the fact that held-out evaluation images are close to the training images. 
For instance, disabling the appearance embeddings ("w/o app") leads to an improvement in PSNR and SSIM. However, Fig.~\ref{fig:qualitative_nerfacto_examples} illustrates that the "w/o app" method results in the production of blurry "floater" artifacts. These artifacts correspond with the training camera locations because the model overfits to small discrepancies in lighting conditions in the training data by placing these artifacts directly in front of the training cameras. (bottom row, bottom left crop). Furthermore, ablations such as "1 prop network" result in subtle changes in the metrics but are more evident in visualizations of the novel views. The use of "1 prop network" as opposed to "Nerfacto (default)" with 2 prop networks leads to aliasing artifacts as can be seen around the small tree branches (bottom row, middle crop). While these artifacts are visible to the eye, especially in the interactive viewer, such temporal discontinuity caused by aliasing is not captured by the quantitative metrics. Furthermore, scene contraction is necessary to correctly recover far objects (top row, right crop). Our conclusions are even more evident in video renderings where the camera moves compared to still image renderings.

Overall, the real-time viewer proves to be useful for viewing out-of-distribution renders. The crops in Fig.~\ref{fig:qualitative_nerfacto_examples} aid in illustrating where certain methods excel over others, regardless of the metrics on the evaluation images. Developing more appropriate evaluation metrics is an important avenue for future research.

\section{Open-source Contributions}

One of the key strengths of our proposed framework is its versatility and ease of use, as demonstrated by our open-source contributions. Our GitHub repository has grown to include over 60 contributors and over 3K stars, reflecting a strong and active community. Additionally, two new libraries, SDFStudio~\cite{Yu2022SDFStudio} and ArcNerf~\cite{tencent_arcnerf}, have been built on top of our framework.
Since the release of Nerfstudio in October 2022, our contributors have enhanced and expanded Nerfstudio by addressing various GitHub issues and feature requests including improved camera paths, colab support, additional camera models, reconstruction of dynamic objects. In the future, we plan on supporting 3D generative pipelines, NeRF compositing, and more.



\section{Conclusion and Future Work}

We draw upon existing techniques and propose a framework that supports a more modularized approach to NeRF development, allows for real-time visualization, and is readily usable with real-world data. We emphasize the importance of utilizing the interactive real-time viewer during training to compensate for imperfect quantitative metrics in model design decisions. We hope the consolidation brought about by this new framework will facilitate the development of NeRF-based methods, thereby accelerating advances in the neural rendering community. Future research directions include the development of more appropriate evaluation metrics and integration of the framework with other fields such as computer vision, computer graphics, and machine learning.





\begin{acks}

This project was supported in part by the Bakar Fellows Program and the BAIR/BDD sponsors. We want to thank the many open-source contributors who have helped create Nerfstudio, including Cyrus Vachha and Rohan Mathur from UC Berkeley and the following Github users:
machenmusik,
kevinddchen (Kevin Chen),
dozeri83,
nikmo33 (Nikhil Mohan),
lsongx (Liangchen Song),
zmurez (Zak Murez),
JulianKnodt (Julian Knodt),
katrinbschmid (Katrin Schmid),
mpmisko (Michal Pandy),
RandomPrototypes,
ManuConcepBrito ,
Zunhammer (Nicolas Zunhammer),
nlml (Liam Schoneveld),
jkulhanek (Jonáš Kulhánek),
mackopes (Martin Piala),
cnsumner,
devernay (Frédéric Devernay),
matsuren (Ren Komatsu),
Mason-McGough (Mason McGough),
hturki,
decrispell (Daniel Crispell),
dkorolov (Dmytro Korolov),
gilureta (Francisca T. Gil Ureta).

\end{acks}

\bibliographystyle{ACM-Reference-Format}
\bibliography{main}

\appendix
\section{Appendix}

This appendix contains the full tables for both the MipNeRF-360~\cite{barron2022mip} comparison (Table~\ref{fig:table-mipnerf360-psnr}) and the Nerfacto ablations (Table~\ref{fig:table-nerfacto-ablations}). Please note that in addition to Nerfacto, we've implemented our own variants of Instant NGP~\cite{muller2022instant}, MipNeRF~\cite{barron2021mip}, Semantic NeRF~\cite{semanticnerf}, the original NeRF~\cite{mildenhall2021nerf}, TensoRF~\cite{chen2022tensorf_tensorf}, and D-NeRF~\cite{pumarola2021d}.


\begin{table*}[h]
\begin{tabular}{|c|ccccccc|}
\hline
\multicolumn{1}{|l}{} & \multicolumn{7}{c|}{$\textbf{PSNR} \uparrow \textbf{for Mip-NeRF 360 Dataset}$} \\ \hline
\textbf{Capture name} & \multicolumn{1}{c|}{\textbf{bicycle}} & \multicolumn{1}{c|}{\textbf{garden}} & \multicolumn{1}{c|}{\textbf{stump}} & \multicolumn{1}{c|}{\textbf{room}} & \multicolumn{1}{c|}{\textbf{counter}} & \multicolumn{1}{c|}{\textbf{kitchen}} & \textbf{bonsai} \\ \hline

NeRF & 21.76 & 23.11 & 21.73 & 28.56 & 25.67 & 26.31 & 26.81 \\
MipNeRF & 21.69 & 23.16 & 23.10 & 28.73 & 25.59 & 26.47 & 27.13 \\
NeRF++ & 22.64 & 24.32 & 24.34 & 28.87 & 26.38 & 27.80 & 29.15 \\
MipNeRF (big MLP) & 22.90 & 25.85 & 23.64 & 30.67 & 28.61 & 29.95 & 31.59 \\
NeRF++ (big MLP) & 23.75 & 25.91 & 25.48 & 30.13 & 27.79 & 29.85 & 30.68 \\
MipNeRF-360 & 24.37 & 26.98 & 26.4 & 31.63 & 29.55 & 32.23 & 33.46 \\
\begin{tabular}[c]{@{}c@{}}Nerfacto (ours) w/o pose \& app\end{tabular} & 24.08 / 22.36 & 26.47 / 24.05 & 24.78 / 18.94 & 30.89 / 29.36 & 27.20 / 25.92 & 30.29 / 28.17 & 32.16 / 28.98\\ \hline


\multicolumn{1}{|l}{} & \multicolumn{7}{c|}{$\textbf{SSIM} \uparrow \textbf{for Mip-NeRF 360 Dataset}$} \\ \hline
\textbf{Capture name} & \multicolumn{1}{c|}{\textbf{bicycle}} & \multicolumn{1}{c|}{\textbf{garden}} & \multicolumn{1}{c|}{\textbf{stump}} & \multicolumn{1}{c|}{\textbf{room}} & \multicolumn{1}{c|}{\textbf{counter}} & \multicolumn{1}{c|}{\textbf{kitchen}} & \textbf{bonsai} \\ \hline

NeRF & 0.455 & 0.546  & 0.453 & 0.843 & 0.755 & 0.749 & 0.792 \\
MipNeRF & 0.454 & 0.543 & 0.517 & 0.851 & 0.779 & 0.745 & 0.818 \\
NeRF++ & 0.526 & 0.635 & 0.594 & 0.852 & 0.802 & 0.816 & 0.876 \\
MipNeRF (big MLP) & 0.612 & 0.777 & 0.643 & 0.903 & 0.877 & 0.902 & 0.928 \\
NeRF++ (big MLP) & 0.630 & 0.761 & 0.687 & 0.883 & 0.857 & 0.888 & 0.913 \\
MipNeRF-360 & 0.685 & 0.813 & 0.744 & 0.913 & 0.894 & 0.920 & 0.941\\
\begin{tabular}[c]{@{}c@{}}Nerfacto (ours) w/o pose \& app\end{tabular} & 0.599 / 0.474 & 0.774 / 0.617 & 0.662 / 0.364 & 0.896 / 0.866 & 0.843 / 0.776 & 0.890 / 0.838 & 0.933 / 0.880\\ \hline


\multicolumn{1}{|l}{} & \multicolumn{7}{c|}{$\textbf{LPIPS} \downarrow \textbf{for Mip-NeRF 360 Dataset}$} \\ \hline
\textbf{Capture name} & \multicolumn{1}{c|}{\textbf{bicycle}} & \multicolumn{1}{c|}{\textbf{garden}} & \multicolumn{1}{c|}{\textbf{stump}} & \multicolumn{1}{c|}{\textbf{room}} & \multicolumn{1}{c|}{\textbf{counter}} & \multicolumn{1}{c|}{\textbf{kitchen}} & \textbf{bonsai} \\ \hline

NeRF & 0.536 & 0.415 & 0.551 & 0.353 & 0.394 & 0.335 & 0.398 \\
MipNeRF & 0.541 & 0.422 & 0.490 & 0.346 & 0.390 & 0.336 & 0.370 \\
NeRF++ & 0.455 & 0.331 & 0.416 & 0.335 & 0.351 & 0.260 & 0.291 \\
MipNeRF (big MLP) & 0.372 & 0.205 & 0.357 & 0.229 & 0.239 & 0.152 & 0.204 \\
NeRF++ (big MLP) & 0.356 & 0.223 & 0.328 & 0.270 & 0.270 & 0.117 & 0.230 \\
MipNeRF-360 & 0.301 & 0.170 & 0.261 & 0.211 & 0.204 & 0.127 & 0.176 \\
\begin{tabular}[c]{@{}c@{}}Nerfacto (ours) w/o pose \& app\end{tabular} & 0.422 / 0.551 & 0.235 / 0.385 & 0.380 / 0.669 & 0.296 / 0.302 & 0.314 / 0.346 & 0.190 / 0.223 & 0.197 / 0.252 \\ \hline

\end{tabular}
\caption{\textbf{PSNR, SSIM, and LPIPS on Mip-NeRF 360 Dataset.} Metrics reported as \{ after 70K iterations ($\sim$30 min) / after 5k iterations ($\sim$5 min) \}.}
\label{fig:table-mipnerf360-psnr}
\vspace{0mm}
\end{table*}

\begin{table*}[h]
\small
\begin{tabular}{|l|lllll}
\hline
 \multicolumn{6}{|c|}{\textbf{PSNR} $\uparrow$ / \textbf{SSIM} $\uparrow$ / \textbf{LPIPS} $\downarrow$ \textbf{for Nerfstudio Dataset}} \\ \hline
Capture name & \multicolumn{1}{l|}{\textbf{Egypt}} & \multicolumn{1}{l|}{\textbf{person}} & \multicolumn{1}{l|}{\textbf{kitchen}} & \multicolumn{1}{l|}{\textbf{plane}} & \multicolumn{1}{l|}{\textbf{dozer}} \\ \hline
Capture device & \multicolumn{1}{l|}{phone} & \multicolumn{1}{l|}{phone} & \multicolumn{1}{l|}{phone} & \multicolumn{1}{l|}{phone} & \multicolumn{1}{l|}{fisheye} \\ \hline
Camera est. method & \multicolumn{1}{l|}{colmap} & \multicolumn{1}{l|}{colmap} & \multicolumn{1}{l|}{polycam} & \multicolumn{1}{l|}{colmap} & \multicolumn{1}{l|}{colmap} \\ \hline


nerfacto & \multicolumn{1}{r}{21.67 / 0.689 / 0.375} & \multicolumn{1}{r}{25.17 / 0.692 / 0.320} & \multicolumn{1}{r}{20.55 / 0.807 / 0.389} & \multicolumn{1}{r}{22.11 / 0.649 / 0.419} & \multicolumn{1}{r|}{20.20 / 0.743 / 0.391} \\ 
w/o pose & \multicolumn{1}{r}{21.56 / 0.689 / 0.374} & \multicolumn{1}{r}{25.52 / 0.716 / 0.323} & \multicolumn{1}{r}{20.47 / 0.798 / 0.402} & \multicolumn{1}{r}{21.83 / 0.645 / 0.421} & \multicolumn{1}{r|}{20.38 / 0.747 / 0.390} \\ 
w/o app & \multicolumn{1}{r}{22.38 / 0.689 / 0.396} & \multicolumn{1}{r}{27.74 / 0.744 / 0.316} & \multicolumn{1}{r}{23.80 / 0.819 / 0.397} & \multicolumn{1}{r}{22.44 / 0.641 / 0.440} & \multicolumn{1}{r|}{23.58 / 0.750 / 0.393} \\ 
w/o pose \& app & \multicolumn{1}{r}{22.31 / 0.692 / 0.399} & \multicolumn{1}{r}{27.76 / 0.749 / 0.328} & \multicolumn{1}{r}{23.39 / 0.808 / 0.409} & \multicolumn{1}{r}{22.22 / 0.642 / 0.443} & \multicolumn{1}{r|}{23.38 / 0.748 / 0.394} \\ 
1 prop network & \multicolumn{1}{r}{21.65 / 0.682 / 0.387} & \multicolumn{1}{r}{25.80 / 0.730 / 0.324} & \multicolumn{1}{r}{20.57 / 0.805 / 0.394} & \multicolumn{1}{r}{22.08 / 0.648 / 0.422} & \multicolumn{1}{r|}{20.00 / 0.738 / 0.403} \\ 
l2 contraction & \multicolumn{1}{r}{21.71 / 0.691 / 0.375} & \multicolumn{1}{r}{25.05 / 0.690 / 0.322} & \multicolumn{1}{r}{20.62 / 0.808 / 0.386} & \multicolumn{1}{r}{22.07 / 0.649 / 0.416} & \multicolumn{1}{r|}{20.47 / 0.745 / 0.386} \\ 
shared prop network & \multicolumn{1}{r}{21.64 / 0.687 / 0.382} & \multicolumn{1}{r}{25.11 / 0.691 / 0.322} & \multicolumn{1}{r}{20.56 / 0.805 / 0.392} & \multicolumn{1}{r}{22.11 / 0.650 / 0.416} & \multicolumn{1}{r|}{20.46 / 0.746 / 0.385} \\ 
random backg. color & \multicolumn{1}{r}{21.56 / 0.678 / 0.389} & \multicolumn{1}{r}{25.54 / 0.709 / 0.328} & \multicolumn{1}{r}{20.64 / 0.809 / 0.390} & \multicolumn{1}{r}{21.95 / 0.642 / 0.423} & \multicolumn{1}{r|}{20.29 / 0.746 / 0.387} \\ 
no contraction & \multicolumn{1}{r}{17.53 / 0.467 / 0.585} & \multicolumn{1}{r}{21.64 / 0.515 / 0.511} & \multicolumn{1}{r}{19.12 / 0.724 / 0.440} & \multicolumn{1}{r}{17.78 / 0.492 / 0.582} & \multicolumn{1}{r|}{18.97 / 0.659 / 0.433} \\ 
synthetic on real & \multicolumn{1}{r}{18.07 / 0.471 / 0.574} & \multicolumn{1}{r}{23.68 / 0.555 / 0.494} & \multicolumn{1}{r}{21.21 / 0.727 / 0.453} & \multicolumn{1}{r}{18.37 / 0.511 / 0.553} & \multicolumn{1}{r|}{21.17 / 0.657 / 0.441} \\ \hline


Capture name & \multicolumn{1}{l|}{\textbf{floating tree}} & \multicolumn{1}{l|}{\textbf{aspen}} & \multicolumn{1}{l|}{\textbf{stump}} & \multicolumn{1}{l|}{\textbf{sculpture}} & \multicolumn{1}{l|}{\textbf{Giannini Hall}} \\ \hline
Capture device & \multicolumn{1}{l|}{fisheye} & \multicolumn{1}{l|}{fisheye} & \multicolumn{1}{l|}{fisheye} & \multicolumn{1}{l|}{fisheye} & \multicolumn{1}{l|}{fisheye} \\ \hline
Camera est. method & \multicolumn{1}{l|}{colmap} & \multicolumn{1}{l|}{colmap} & \multicolumn{1}{l|}{colmap} & \multicolumn{1}{l|}{colmap} & \multicolumn{1}{l|}{colmap} \\ \hline


nerfacto & \multicolumn{1}{r}{20.03 / 0.740 / 0.312} & \multicolumn{1}{r}{16.95 / 0.347 / 0.560} & \multicolumn{1}{r}{22.04 / 0.698 / 0.295} & \multicolumn{1}{r}{21.88 / 0.676 / 0.368} & \multicolumn{1}{r|}{19.26 / 0.593 / 0.462} \\ 
w/o pose & \multicolumn{1}{r}{19.80 / 0.721 / 0.321} & \multicolumn{1}{r}{16.96 / 0.345 / 0.560} & \multicolumn{1}{r}{21.65 / 0.667 / 0.305} & \multicolumn{1}{r}{21.82 / 0.670 / 0.368} & \multicolumn{1}{r|}{19.28 / 0.591 / 0.461} \\ 
w/o app & \multicolumn{1}{r}{21.84 / 0.727 / 0.329} & \multicolumn{1}{r}{16.92 / 0.333 / 0.567} & \multicolumn{1}{r}{25.11 / 0.777 / 0.308} & \multicolumn{1}{r}{22.16 / 0.657 / 0.422} & \multicolumn{1}{r|}{20.54 / 0.587 / 0.495} \\ 
w/o pose \& app & \multicolumn{1}{r}{21.77 / 0.719 / 0.341} & \multicolumn{1}{r}{16.91 / 0.334 / 0.567} & \multicolumn{1}{r}{24.87 / 0.778 / 0.307} & \multicolumn{1}{r}{22.08 / 0.650 / 0.428} & \multicolumn{1}{r|}{20.63 / 0.587 / 0.494} \\ 
1 prop network & \multicolumn{1}{r}{19.96 / 0.736 / 0.318} & \multicolumn{1}{r}{16.79 / 0.325 / 0.587} & \multicolumn{1}{r}{22.70 / 0.751 / 0.283} & \multicolumn{1}{r}{21.99 / 0.685 / 0.365} & \multicolumn{1}{r|}{19.20 / 0.585 / 0.475} \\ 
l2 contraction & \multicolumn{1}{r}{19.81 / 0.741 / 0.311} & \multicolumn{1}{r}{16.89 / 0.346 / 0.562} & \multicolumn{1}{r}{22.16 / 0.708 / 0.296} & \multicolumn{1}{r}{21.84 / 0.673 / 0.369} & \multicolumn{1}{r|}{19.21 / 0.592 / 0.457} \\ 
shared prop network & \multicolumn{1}{r}{19.91 / 0.734 / 0.315} & \multicolumn{1}{r}{16.95 / 0.344 / 0.561} & \multicolumn{1}{r}{21.73 / 0.692 / 0.301} & \multicolumn{1}{r}{21.75 / 0.670 / 0.374} & \multicolumn{1}{r|}{19.25 / 0.590 / 0.462} \\ 
random backg. color & \multicolumn{1}{r}{19.85 / 0.729 / 0.316} & \multicolumn{1}{r}{16.95 / 0.343 / 0.561} & \multicolumn{1}{r}{22.20 / 0.711 / 0.294} & \multicolumn{1}{r}{21.84 / 0.677 / 0.370} & \multicolumn{1}{r|}{19.23 / 0.590 / 0.457} \\ 
no contraction & \multicolumn{1}{r}{18.85 / 0.588 / 0.410} & \multicolumn{1}{r}{14.13 / 0.170 / 0.717} & \multicolumn{1}{r}{20.27 / 0.630 / 0.392} & \multicolumn{1}{r}{20.71 / 0.619 / 0.436} & \multicolumn{1}{r|}{16.92 / 0.472 / 0.557} \\ 
synthetic on real & \multicolumn{1}{r}{20.64 / 0.565 / 0.418} & \multicolumn{1}{r}{14.76 / 0.186 / 0.690} & \multicolumn{1}{r}{23.56 / 0.675 / 0.400} & \multicolumn{1}{r}{21.71 / 0.609 / 0.485} & \multicolumn{1}{r|}{17.74 / 0.468 / 0.579} \\ \hline


\end{tabular}
\caption{\textbf{Ablations of Nerfacto method on Nerfstudio Dataset.} We remove and change various components of the Nerfacto method and report \{ PSNR, SSIM, LPIPS \} on the Nerfstudio Dataset. "synthetic on real" is our synthetic settings adjusted to work on a real-world capture (i.e., a bounded scene, no appearance embeddings, and no scene contraction). For the "1 prop network" experiment, we use only the first proposal network from Nerfacto and only do one iteration of the proposal sampling instead of two.}
\label{fig:table-nerfacto-ablations}
\vspace{0mm}
\end{table*}

\end{document}